\documentclass[10pt,twocolumn,oneside,letterpaper,conference]{IEEEtran}
\IEEEoverridecommandlockouts
\usepackage{cite}
\usepackage{amsmath,amssymb,amsfonts}
\usepackage{graphicx}
\usepackage{textcomp}
\usepackage{xcolor}
\usepackage{xcolor}

\usepackage{cite}
\usepackage{graphicx}
\usepackage{epstopdf}

\usepackage{amsmath}
\usepackage{times}
\usepackage{latexsym}
\usepackage{graphicx}
\usepackage{bm}
\usepackage{amssymb}
\usepackage{caption}
\usepackage{stfloats}
\usepackage{cases}
\usepackage{array}
\usepackage{setspace}
\usepackage{fancyhdr}
\usepackage{fancyhdr}
\usepackage{algorithm}
\usepackage{algorithmicx}
\usepackage{algpseudocode}
\usepackage{textcomp}
\usepackage{wasysym}
\usepackage{balance} 
\usepackage{url}
\usepackage{xcolor}
\usepackage{subfigure}
\usepackage{stfloats}
\usepackage{enumerate}
\usepackage{array}
\usepackage{xcolor}
\usepackage{tikz}
\usepackage{booktabs}
\usepackage{colortbl}
\usepackage{verbatim}
\usepackage{balance}

\def\BibTeX{{\rm B\kern-.05em{\sc i\kern-.025em b}\kern-.08em
   T\kern-.1667em\lower.7ex\hbox{E}\kern-.125emX}}
\begin{document}

\title{Computation Offloading and Resource Allocation in F-RANs: A Federated Deep Reinforcement Learning Approach
}

\author{
\IEEEauthorblockN{Lingling~Zhang$^{1}$,
Yanxiang Jiang$^{1,2,*}$, Fu-Chun Zheng$^{1,2}$, Mehdi Bennis$^3$,
and Xiaohu You$^1$}
\IEEEauthorblockA{$^1$National Mobile Communications Research Laboratory,
Southeast University, Nanjing 210096, China\\
$^2$School of Electronic and Information Engineering, Harbin Institute of Technology, Shenzhen 518055, China\\
$^3$Centre for Wireless Communications, University of Oulu, Oulu 90014, Finland\\
E-mail: $\{$zhangllling@seu.edu.cn, yxjiang@seu.edu.cn, fzheng@ieee.org, mehdi.bennis@oulu.fi, xhyu@seu.edu.cn$\}$
}}
\maketitle

\begin{abstract}
The fog radio access network (F-RAN) is a promising technology in which the user mobile devices (MDs) can offload computation tasks to the nearby fog access points (F-APs). Due to the limited resource of F-APs, it is important to design an efficient task offloading scheme. In this paper, by considering time-varying network environment, a dynamic computation offloading and resource allocation problem in F-RANs is formulated to minimize the task execution delay and energy consumption of MDs. To solve the problem, a federated deep reinforcement learning (DRL) based algorithm is proposed, where the deep deterministic policy gradient (DDPG) algorithm performs computation offloading and resource allocation in each F-AP. Federated learning is exploited to train the DDPG agents in order to decrease the computing complexity of training process and protect the user privacy. Simulation results show that the proposed federated DDPG algorithm can achieve lower task execution delay and energy consumption of MDs more quickly compared with the other existing strategies.

\end{abstract}

\section{Introduction}
As a large number of intelligent devices access the wireless network, the traffic of wireless data is growing explosively, which brings great challenges to the cloud radio access network (C-RAN)\cite{7880686}\cite{8937820}. The fog radio access network (F-RAN) is thus proposed, which places computing, storage, and network resources in fog access points (F-APs) that are closer to the users\cite{7513863}\cite{9013376}\cite{9200664}. Users can offload tasks to the F-APs via wireless links to support computation intensive and delay sensitive applications on their mobile devices (MDs), thus to cope with the computation capability and available energy constraints\cite{7792373}. However, the F-APs may not be able to serve all users at the same time due to the limited computation and communication capabilities. In this case, the computation offloading strategy needs to be carefully designed to improve the Quality of Service (QoS) for users.

The computation offloading problem has been investigated in many research works. In \cite{8334188}, the binary computation offloading problem for multi-users was solved by an alternating direction method of multipliers (ADMM). The game theory was used to deal with the joint problem of computation offloading and resource allocation in \cite{8985335}\cite{8756083}. Reference \cite{8533343} adopted the convex/quasi convex optimization and meta heuristic algorithm to solve the sub problems of resource allocation and offloading decision respectively. However, these optimization-based methods are time consuming because of the high computational complexity, and not applicable to real scenarios. To deal with the complex and dynamic scenes, the machine learning based approaches are increasingly employed, especially deep reinforcement learning (DRL) methods. In \cite{8771176}, an online algorithm based on DRL was designed under the time-varying channel conditions. Researches in \cite{9290426} and \cite{9205252} applied deep Q network (DQN) to solve the computation offloading problem. But these value-based DRL methods can only deal with discrete action spaces and need discretization operations when the action is continuous. To handle continuous action spaces, reference \cite{9288861} combined deep neuroevolution and policy gradient to solve the computation offloading problem. The authors in \cite{9039641} \cite{9158401} \cite{9310745} exploited deep deterministic policy gradient (DDPG) algorithm to perform computation offloading. In these works, centralized methods are mostly used to train the DRL agents, which will cause exponential growth of the computing complexity and system overhead with the increase of the number of users. The decentralized training method in \cite{9310745} does not have this trouble, but it does not take the cooperation among agents into account thus the global system performance is not as good as that of centralized methods.

In this paper, by considering the dynamic environment as well as the continuous action of resource allocation, a DDPG algorithm for computation offloading and resource allocation in F-RANs is proposed. Further, a federated DDPG algorithm which applies the federated training method among the DDPG agents is proposed to decrease the computing complexity, improve the overall performance and protect user privacy. The rest of the paper is organized as follows. In Section II, the system model is presented and the problem of computation offloading and resource allocation is formulated. The proposed federated DRL algorithm is described in Section III and the algorithm performance is evaluated in Section IV. Finally, conclusions are drawn in Section V.

\section{System Model and Problem Formulation}
\subsection{System Model}
The considered network architecture of the F-RAN is shown in Fig. 1. The F-RAN consists of  a cloud center, a set $\mathsf{\mathcal{N}}\text{=}\left\{ 1,2,...,N \right\}$ of  F-APs, and a set $\mathsf{\mathcal{M}}\text{=}\left\{ 1,2,...,M \right\}$ of MDs within the coverage of F-AP $n$ ($n \in \mathcal{N}$). Each F-AP is equipped with multiple antennas, while each MD has a single antenna and connects to the F-AP via wireless channels. Note that the number of antennas in the F-AP is larger than the number of MDs. A discrete-time model is applied in the F-RAN system, where the time is divided into slots with equal length ${{T}_{s}}$ (in seconds) and indexed by $\mathsf{\mathcal{T}}\text{=}\left\{ 0,1,...,T-1 \right\}$.

At time slot $t$, each MD has a computation intensive task to complete. The task generated by MD $m$ within F-AP $n$ can be described as $({{b}_{m}^{n}}(t)$,${{d}_{m}^{n}}(t))$, where ${{b}_{m}^{n}}(t)$ denotes the task size (in bits) and ${{d}_{m}^{n}}(t)$ denotes the required computation resource (i.e., CPU cycles) for task completion. The task can be processed locally by the MD, or offloaded to the corresponding F-AP. Note that the binary offloading mode is considered in this paper.

\begin{figure}[t]
\centering 
\includegraphics[height=4.0cm,width=7.2cm]{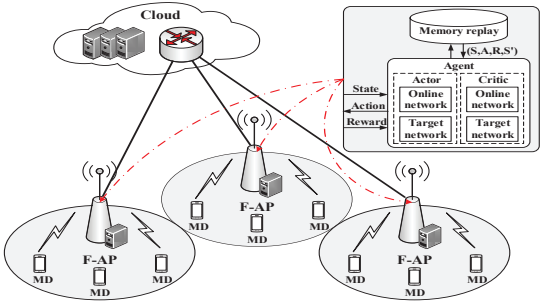}
\caption{Architecture of the F-RAN.}
\label{scenario}
\end{figure}\

\subsection{Computing Models}
\subsubsection{Local computing}
Let ${{x}_{m}^{n}}(t) \in \left\{ 0,1 \right\}$ denote the offloading decision variable of MD $m$ within F-AP $n$. Specifically, ${{x}_{m}^{n}(t)}=0$ represents that the task generated by MD $m$ within F-AP $n$ at slot $t$ is processed locally, otherwise the task is offloaded to the F-AP $n$ via a wireless link. Define ${f}_{m}^{n}$ as the CPU frequency of MD $m$ within F-AP $n$. If the task is decided to execute on the MD $m$, the local computing delay can be calculated as follows
\begin{equation}
	{{t}_{mn}^{l}}\left( t \right)=\frac{{{d}_{m}^{n}}(t)}{{f}_{m}^{n}}\text{.}
\end{equation}
The corresponding energy consumption of the MD is
\begin{equation}
	{{e}_{mn}^{l}}\left( t \right)={{\xi}_{m}^{n}}{{d}_{m}^{n}}(t)\text{,}
\end{equation}
where ${\xi}_{m}^{n}$ denotes the energy consumption coefficient per CPU cycle of MD $m$, which is related to the chip architecture. Here, we set ${\xi}_{m}^{n}\text{=}10^{-27}{\left({f}_{m}^{n}\right)}^{2}$ according to \cite{6195685}.

\subsubsection{F-AP computing}
Define $f_n$ as the CPU frequency of F-AP $n$. Since the computation capability of F-APs is limited, the MDs that offload tasks to the corresponding F-AP can only be allocated partial computation resource. Let ${{y}_{m}^{n}}(t) \in \left[ 0,1 \right]$ denote the computation resource allocation ratio of MD $m$ from F-AP $n$. If the task of MD $m$ is offloaded to F-AP $n$ for completion, the F-AP computing delay can be given by
\begin{equation}
	{{t}_{mn}^{o}}\left( t \right)=\frac{{{d}_{m}^{n}}(t)}{{y}_{m}^{n}(t){f_n}}\text{.}
\end{equation}

\subsection{Communication Model}
If computation tasks are executed on the F-APs, the task data and computation results need to be transmitted between the MDs and corresponding F-APs. Since the size of computing results is much smaller than that of task data, the delay and energy consumption of result transmission from F-APs to MDs are ignored in this work.

An orthogonal frequency division multiple access (OFDMA) protocol is adopted in wireless networks, in which F-AP $n$ provides a total bandwidth $B_n$ for covered MDs. Define ${{z}_{m}^{n}}(t) \in \left[ 0,1 \right]$ as the bandwidth allocation ratio of MD $m$ from F-AP $n$, then the achievable uplink data rate of MD $m$ can be calculated by
\begin{equation}
	{{r}_{m}^{n}}\left( t \right)={z}_{m}^{n}\left( t \right){B_n}{\log}_{2}(1+\frac{{p_m^n}{{g}_{m}^{n}}(t)}{{\sigma}^{2}})\text{,}
\end{equation}
where ${\sigma}^{2}$ is the background noise power, $p_m^n$ is the transmit power of MD $m$,  and ${g}_{m}^{n}(t)$ is the channel gain between MD $m$ and F-AP $n$, which is related to the distance between them. It is worth noting that the MDs are fixed in a time slot and move randomly between slots. When MD $m$ offloads computation task to F-AP $n$, the transmission delay can be expressed as
\begin{equation}
	{{t}_{mn}^{tr}}\left( t \right)=\frac{{{b}_{m}^{n}}(t)}{{r}_{m}^{n}(t)}\text{.}
\end{equation}
The corresponding energy consumption of MD $m$ for transmission can be calculated by
\begin{equation}
	{{e}_{mn}^{tr}}\left( t \right)={{p}_{m}^{n}}{{t}_{mn}^{tr}}(t)\text{.}
\end{equation}

Based on the F-AP computing and communication models, when the computation task of MD $m$ is offloaded to F-AP $n$, the total execution delay includes computation delay and transmission delay, which can be written as
\begin{equation}
	{{t}_{mn}^{e}}\left( t \right)={{t}_{mn}^{o}}(t) + {{t}_{mn}^{tr}}(t)\text{.}
\end{equation}

\subsection{Problem Formulation}
Given the offloading decision $x_m^n(t)$ of MD $m$ at time slot $t$, the total computation delay and energy consumption of all MDs covered by F-AP $n$ can be calculated respectively as
\begin{equation}
	{T}_{n}\left( t \right)=\sum_{m\in \mathcal{M}}{(1-x_m^n(t)){t_{mn}^{l}}(t)+x_m^n(t){t_{mn}^e(t)}}  \text{,}
\end{equation}\
\begin{equation}
	{E}_{n}\left( t \right)=\sum_{m\in \mathcal{M}}{(1-x_m^n(t)){e_{mn}^{l}}(t)+x_m^n(t){e_{mn}^{tr}(t)}}  \text{.}
\end{equation}
In terms of task execution delay and energy consumption, the overall cost of all MDs within F-AP $n$ at slot $t$ is defined as
\begin{equation}
	{C}_{n}\left( t \right)=\omega T_n(t)+\varpi E_n(t) \text{,}
\end{equation}
where $\omega$ and $\varpi$ are weigh parameters that control the tradeoff among delay and energy cost and satisfy $\omega+\varpi\text{=}1$.

The objective of this paper is to minimize the long term average computation delay and energy consumption of all MDs in F-RANs, thus the computation offloading and resource allocation problem can be formulated as follows:
\begin{align}\label{P-1}
{\textbf P: }  &\underset{\mathbf{x}\text{,}\mathbf{y}\text{,}\mathbf{z}}{\mathop{\min }}\,{\lim_{T \to \infty}\frac{1}{T} \sum_{t\in \mathcal{T}}{\mathbb{E}\left[ \sum_{n\in \mathcal{N}}{C_n \left( t \right)}\right]}} \\
{\text{s.t.}}\ &x_m^n(t)\in \left\{ 0,1 \right\}\text{,}\forall m\text{,}n\text{,}t\text{,} \tag{\ref{P-1}a}\nonumber \label{con_a} \\
&y_m^n(t)\in \left[ 0,1 \right]\text{,}\forall m\text{,}n\text{,}t\text{,} \tag{\ref{P-1}b}\nonumber \label{con_b} \\
&z_m^n(t)\in \left[ 0,1 \right]\text{,}\forall m\text{,}n\text{,}t\text{,}  \tag{\ref{P-1}c}\nonumber \label{con_c} \\
&\sum_{m\in \mathcal{M}}{y_m^n(t) \le 1}\text{,}\forall n\text{,}t\text{,}  \tag{\ref{P-1}d}\nonumber \label{con_d} \\
&\sum_{m\in \mathcal{M}}{z_m^n(t) \le 1}\text{,}\forall n\text{,}t\text{,} \tag{\ref{P-1}e}\nonumber \label{con_e}
\end{align}
where $\mathbf{x}\text{=}\left\{x_m^n(t)\right\}$, $\mathbf{y}\text{=}\left\{y_m^n(t)\right\}$ and $\mathbf{z}\text{=}\left\{z_m^n(t)\right\}$. Constraints (11d) and (11e) represent that the sum of computation resource and communication resource allocated to MDs cannot exceed the capabilities of the corresponding F-AP, respectively.

The optimization problem \textbf{P} in (11) is a mixed integer non-linear programming problem (MINLP), in which the offloading decision indicator $x_m^n(t)$ is a binary variable, while the computation resource allocation ratio $y_m^n(t)$ and communication resource allocation ratio $z_m^n(t)$ are continuous variables. Besides, problem \textbf{P} is a non-deterministic polynomial (NP)-hard problem, whose complexity will significantly increase as the number of MDs and F-APs increases. Adopting traditional optimization based methods to solve the problem is time-consuming and inflexible, and not suitable for time-varying environments in F-RANs. As DRL algorithms among intelligent learning algorithms are able to use less information to perceive the environment and make decisions on complex problems, we exploit DRL to solve the optimization problem.

\section{Proposed Federated Deep Reinforcement Learning Algorithm}
In order to solve the problem \textbf{P} via a DRL based method, we first reformulate the problem in the RL framework and then propose a federated DRL-based algorithm.
\subsection{Problem Transformation With DRL}
In a RL approach, an agent continuously interacts with the environment and learns to better adapt to the environment. The RL agent usually makes decisions based on the Markov Decision Process (MDP). A typical MDP is defined by a 4-tuple $(\mathcal{S}, \mathcal{A}, \mathcal{P}, \mathcal{R})$, which consists of a set of possible states $\mathcal{S}$, a set of available actions $\mathcal{A}$, a reward function $\mathcal{R}$ and transition probability $\mathcal{P}$. At an arbitrary step $t$, the RL agent observes the environment state $s_t \in \mathcal{S}$, and takes an action $a_t \in \mathcal{A}$ according to the specific policy $\pi$. Then the agent will receive a reward $r_t\text{=}\mathcal{R}(s_t, a_t)$ from the environment and the state will transit from $s_t$ to $s_{t+1}$. Next, the agent will continue to take a new action based on the state $s_{t+1}$ and get a new reward. The learning process will continue iteratively in this way, aiming to maximize the long-term discounted reward of the RL agent, i.e., $R\text{=}\sum_{t=0}^{\infty}{{\gamma ^t} r_t}$ , where $\gamma \in \left[0,1\right]$ is a discount factor.

The goal of the RL agent is to learn the optimal policy to solve the MDP. As the action-value function $Q^{\pi}(s,a)$ represents the expected reward under the policy $\pi$, the optimal policy ${\pi}^*$ which chooses the optimal action greedily in state $s$ is expressed as
\begin{equation}
	\pi^*(s)=\text{arg}\underset{a}{\mathop{\max }}\,{Q^*(s\text{,}a)} \text{.}
\end{equation}

In order to exploit DRL to solve the computation offloading and resource allocation problem, we transform the problem \textbf{P} as MDP form as follows.
\subsubsection{State}
At time slot $t$, the MDs send information of their computation tasks and location to the corresponding F-AP. Each F-AP receives these information and calculate the channel gains between itself and MDs. Therefore, the system state of F-AP $n$ at slot $t$ can be described as
\begin{equation}
	\boldsymbol{S}_n(t)=\left\{\boldsymbol{B}_n(t)\text{,}\boldsymbol{D}_n(t)\text{,}L_n^{F-AP}\text{,}\boldsymbol{L}_n^{MD}(t)\text{,}\boldsymbol{G}_n(t)\right\} \text{,}
\end{equation}
where $\boldsymbol{B}_n(t)=\left\{ b_m^n(t)\text{,} m \in \mathcal{M} \right\}$ is the task size vector, $\boldsymbol{D}_n(t)=\left\{ d_m^n(t)\text{,}m \in \mathcal{M} \right\}$ is the task computation resource vector, $L_n^{F-AP}$ is the location of F-AP $n$, $\boldsymbol{L}_n^{MD}(t)$ is the location of MDs within F-AP $n$, and $\boldsymbol{G}_n(t)=\left\{ g_m^n(t)\text{,}m \in \mathcal{M} \right\}$ is the channel gain vector.
\subsubsection{Action}
In each time slot $t$, the F-AP $n$ generates an action based on the current state. The action includes three parts: computation offloading decision $\boldsymbol{x}_n(t)=\left\{ x_m^n(t)\text{,}m \in \mathcal{M} \right\}$, the corresponding computation resource allocation vector $\boldsymbol{y}_n(t)=\left\{ y_m^n(t)\text{,}m \in \mathcal{M} \right\}$ and communication resource allocation vector $\boldsymbol{z}_n(t)=\left\{ z_m^n(t)\text{,}m \in \mathcal{M} \right\}$ of all MDs within F-AP $n$.
\subsubsection{Reward}
The objective of this work is to minimize the delay and energy consumption of MDs. Given the particular state $s_t$ and action $a_t$ at time slot $t$, the immediate reward function of F-AP $n$ can be expressed by $C_n(t)$ in (10). Since the objective of DRL is to maximize the reward, the long-term reward value should be negatively correlated to the immediate reward, which is expressed as
\begin{equation}
	R_n(t)=-\sum_{t=0}^{\infty}{\gamma^t{C_n(t)}} \text{.}
\end{equation}
\subsection{DDPG-Based Computation Offloading and Resource Allocation Algorithm}
As the environment state dynamically changes over time in F-RANs, a model-free DRL algorithm is an excellent choice for problem \textbf{P}. Since the resource allocation action is continuous in this work, a DDPG algorithm is exploited to solve the computation offloading and resource allocation problem. To avoid the MDP becomes more complex with the increase of number of F-APs and MDs, we propose to deploy DDPG agent at each F-AP as shown in Fig. 1 and train them with federated learning (FL) approach. After observing the environment state, each agent finds optimal actions and sends the offloading decisions to the MDs. Once the information of optimal actions is received, the MDs execute the computation tasks locally or transmit them to the F-AP for completion with the allocated communication and computation resource. The process of DDPG-based computation offloading and resource allocation algorithm is described as follows.

DDPG algorithm is based on an actor-critic architecture that adopts two separate deep neural networks (DNNs), where actor is responsible for policy network and critic is responsible for value network. At the beginning, the F-AP agent obtains the environment information and formulates the state space. Then the actor network chooses an action $a_t$ according to the current state and policy and adds stochastic noise
\begin{equation}
	a_t={\pi}(s_t \text{;} \boldsymbol{\theta}^\pi)+\zeta \text{,}
\end{equation}
where $\zeta$ is Gaussian noise. After taking the action $a_t$, the environment returns a reward $r_t \text{=}-C_n(t)$ and new state $s_{t+1}$. The F-AP stores the signal $(s_t, a_t, r_t, s_{t+1})$ generated in the state transition process into the memory replay $\mathcal{B}$. Next, the F-AP picks $K$ samples randomly from the memory replay $\mathcal{B}$ as the training sets for the online actor network and online critic network. The critic network is updated by minimizing the loss function, which is expressed as
\begin{equation}
	L=\frac{1}{K}\sum_{i}{(y_i-Q(s_i\text{,}a_i \text{;} \boldsymbol{\theta}^Q))^2} \text{,}
\end{equation}
where $y_i$ represents the target Q value
\begin{equation}
	y_i=r_i+\gamma Q'(s_{i+1}\text{,}\pi{'}(s_{i+1}\text{;}\boldsymbol{\theta}^{\pi^{'}})\text{;}\boldsymbol{\theta}^{Q^{'}}) \text{.}
\end{equation}
The actor network is updated by using stochastic gradient descent method as
\begin{equation}
	\nabla_{\boldsymbol{\theta}^{\pi}}J\approx \frac{1}{K} \sum_{i}\nabla_{a}Q(s\text{,}a\text{;}\boldsymbol{\theta}^Q)|_{s=s_i,a=\pi(s_i)}\nabla_{\boldsymbol{\theta}^{\pi}}\pi(s\text{;}\boldsymbol{\theta}^{\pi})|_{s_i} \text{.}
\end{equation}
Lastly, a soft updating method is exploited to update the parameters of target networks as
\begin{equation}
	\boldsymbol{\theta}^{{Q}^{\prime}} \gets \tau \boldsymbol\theta^Q + (1-\tau)  \boldsymbol{\theta}^{{Q}^{\prime}} \text{,}
\end{equation}
\begin{equation}
    \boldsymbol{\theta}^{{\pi}^{\prime}} \gets \tau \boldsymbol\theta^\pi + (1-\tau)  \boldsymbol{\theta}^{{\pi}^{\prime}} \text{.}
\end{equation}
where $\tau$ is the soft update coefficient. Details of the proposed DDPG solution are described in Algorithm 1.
\begin{algorithm}[!t]
	\label{alg:1}
	\begin{algorithmic}[1]
		\caption{DDPG-Based Computation Offloading and Resource Allocation Algorithm per F-AP}
		\begin{spacing}{0.8}
			\item \textbf{Inputs:} System model parameters, number of episodes, number of time steps, learning rates of actor and critic networks, reward discount factor, memory size, sample size.
            \item \textbf{Initialization:}
			\item  Initialize the weights $\boldsymbol\theta^Q$,$\boldsymbol\theta^\pi$ of the critic online network $Q(s\text{,}a \text{;} \boldsymbol{\theta}^Q)$ and actor online network $\pi(s\text{;}\boldsymbol{\theta}^{\pi})$;
			\item Initialize the target networks $Q'$ and $\pi'$ with weights $\boldsymbol{\theta}^{{Q}^{\prime}} \gets \boldsymbol\theta^Q$,$\boldsymbol{\theta}^{{\pi}^{\prime}} \gets \boldsymbol\theta^\pi$;
            \item Initialize the memory replay $\mathcal{B}$.
			\For {episode $k$=1,2,...,$k_{max}$}
				\State Generate an initial state $s_0$;
				\For {time step $t=0,1,...,T-1$}
					\State Based on the current policy with Gaussian \hspace*{2.0\dimexpr\algorithmicindent}noise, choose action $a_t$ according to (15);	
					\State Execute action $a_t$ and record the reward $r_t$ and \hspace*{2.0\dimexpr\algorithmicindent}new state $s_{t+1}$;
					\State Store the transition $(s_t,a_t,r_t,s_{t+1})$ in the memory \hspace*{2.0\dimexpr\algorithmicindent}replay $\mathcal{B}$. If $\mathcal{B}$ is full, replace the oldest ones;
                    \State Randomly sample mini-batch $K$ of transition \hspace*{2.0\dimexpr\algorithmicindent}tuples $(s_t,a_t,r_t,s_{t+1})$ from $\mathcal{B}$;
                    \State Calculate target Q value $y_i$ by (17);
                    \State Update critic online network according to (16);
                    \State Update actor online network according to (18);
                    \State Update target networks according to (19) and (20);
				\EndFor
			\EndFor
            \item \textbf{Output:} Optimal policy $\pi^*$.
		\end{spacing}
	\end{algorithmic}
	\vspace{-1.0em}
\end{algorithm}
\subsection{Federated DDPG Algorithm}
In the centralized training method, the cloud center is considered as the DRL agent. The MDs served by all the F-APs send their computation requests and related information to the cloud. The cloud trains the DRL model to find the optimal actions for all MDs and then informs them. This workflow brings significant challenges. On the one hand, the cloud needs to keep a large scale model that serves for all MDs, which will lead to great computing complexity and consume a great deal of computation and storage resource. On the other hand, the direct communication between the cloud and MDs or indirect communication through F-APs to deliver large amounts of user data will cause heavy burden of the network. In addition, for privacy reasons, the users may be unwilling to transmit the offloading requests and information to the cloud.

To this end, we exploit FL to distributively train the DDPG agents located at the F-APs. The proposed federated DDPG algorithm is described in Algorithm 2. Specifically, at each round $j$ of the training process, the F-APs download the global DDPG weights $H(j)$ from the cloud center. Then each F-AP trains the DDPG agent locally by the local data and transmit the updated local model weights $H_n(j) (n \in \mathcal{N})$ to the cloud. The cloud aggregates the received parameters by federated averaging as follows to obtain an updated global model,
\begin{equation}
    H(j+1)=\frac{1}{N} \sum_{n=1}^N H_n(j) \text{.}
\end{equation}
After that, the cloud distributes the updated global model weights to the F-APs thus to conduct the training process on F-APs using the new parameters. The iteration will repeat until the model converges.
\begin{algorithm}[!t]
	\label{alg:1}
	\begin{algorithmic}[1]
		\caption{Federated DDPG Algorithm for Computation Offloading and Resource Allocation}
		\begin{spacing}{0.8}
            \item \textbf{Initialization:}
			\item Initialize the global DRL agent with random weights $H(0)$ at the cloud side;
            \item Initialize the local DRL model weights $H_n(0) (n \in \mathcal{N})$ at the F-APs side;
            \item F-APs download $H(0)$ from the cloud and set $H_n(0)\text{=} H(0)$;
            \item Initialize the memory replay $\mathcal{B}$ of F-APs.
            \item \textbf{Iteration:}
			\For {round $j=1, 2, ... , j_{max}$}
				\For {each F-AP $n \in \mathcal{N}$ in parallel}
					\State Download $H(j)$ from the cloud and set \hspace*{2.0\dimexpr\algorithmicindent}$H_n(j)\text{=} H(j)$;	
					\State Train the DRL agent locally with $H_n(j)$;
					\State Upload the trained weights $H_n(j+1)$ to the cloud;
                \EndFor
                \State With respect to the cloud:
                \State \hspace*{1.0\dimexpr\algorithmicindent}Collect all weights $H_n(j)$ updates;
                \State \hspace*{1.0\dimexpr\algorithmicindent}Conduct federated averaging;
                \State \hspace*{1.0\dimexpr\algorithmicindent}Broadcast averaged weights $H_n(j+1)$;
			\EndFor
		\end{spacing}
	\end{algorithmic}
	\vspace{-1.0em}
\end{algorithm}

\section{Simulation Results}
\subsection{Simulation Setup}
We consider a network of $N=4$ F-APs, each of which covers an area of 200 m $\times$ 200 m and provides wireless services with a bandwidth of 10 MHz for $M=5$ randomly distributed MDs. According to \cite{2002Wireless}, the channel gain is set as $g_m^n=d_{m\text{,}n}^{-\alpha}$, where $d_{m\text{,}n}$ is the distance between MD $m$ and F-AP $n$ and $\alpha=4$ is the pass loss factor. The transmission power of each MD is uniformly distributed between 0.1 W and 1 W, and the background noise $\sigma^2$= -100 dBm. For the computation task, the data size is randomly distributed in the range from 200 KB to 300 KB, and the required CPU cycles per bit is in the range from 200 to 500. The CPU computation capability of each MD is randomly distributed between 1 GHz and 2 GHz, while that of F-APs is 5 GHz. In addition, the weigh factors of delay and energy are set as $\omega=\varpi=0.5$ and the duration of a time slot is $T_s$=1 s.

The actor and critic of the DDPG agent in the simulation are four-layer fully connected neural networks with two hidden layers, each consisting of 300 and 100 neurons respectively. The activation function of the hidden layers is rectified linear unit (ReLU), and that of the output layer in actor network is sigmoid function. The neural network parameters are updated by the Adam optimizer, in which the learning rates of actor and critic are 0.001 and 0.0001, respectively. Replay memory capacity is set to 20000, and the mini-batch size is 64. The discount factor is set to 0.9, and the soft updating rate of target networks is set to 0.001.

To verify the performance of the federated DDPG algorithm, several benchmark algorithms are introduced as follows.
\subsubsection{DQN algorithm}
The F-APs adopt DQN algorithm whose neural network architecture and parameters are the same as that of DDPG. To handle continuous actions, the action spaces for computation and communication resource allocation are both discretized uniformly into 5 levels each.
\subsubsection{Local computing}
Each MD executes the generated computation tasks locally.
\subsubsection{F-AP computing}
All tasks are offloaded to the corresponding F-APs for completion. The F-APs equally allocate the computation and communication resource to MDs.
\subsection{Results and Analysis}
The convergence performance of the proposed federated DDPG algorithm is compared with the centralized DDPG and DQN algorithms in Fig. 2, where the vertical axis is the average system reward of all F-APs and the horizontal axis is the training episode. Some points can be obtained from the Fig. 2. First, the federated DRL algorithms are superior to the corresponding centralized DRL algorithms, which converge faster to higher system rewards. This is because centralized DRL algorithms have much larger action spaces to handle than federated DRL algorithms. Second, the DDPG algorithms converge faster with better stability compared with the DQN algorithms, where the convergence speed of federated DDPG algorithm is four times that of federated DQN algorithm. The reason is that DDPG directly outputs determined action without processing a large amount of discrete actions like DQN. Third, the federated DDPG algorithm outperforms other algorithms, with the fastest convergence speed and the highest system reward. Therefore, the federated DDPG algorithm responds to users more quickly with lower task computation delay and energy consumption than that of other three algorithms, which demonstrates the effectiveness and efficiency of the proposed algorithm.
\begin{figure}[t]
	\centering 
	\includegraphics[height=4.5cm,width=7.5cm]{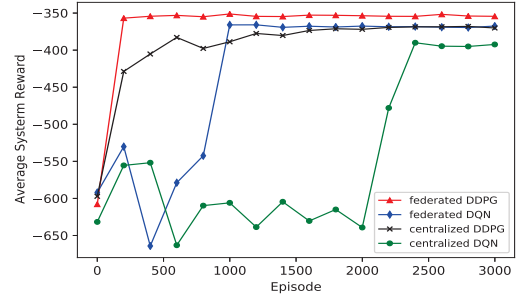}
	\caption{Convergence performance of different algorithms.}
	\label{1}
\end{figure}\

The average system costs (i.e., the weighted sum of delay and energy in (10)) of various algorithms under different number of MDs in each F-AP are shown in Fig. 3. We can see that the average system costs of all algorithms are positively correlated with the number of MDs. As the MDs in each F-AP increases, more computation tasks are generated and thus results in the increase of the computation delay and energy consumption. Among all algorithms, the average system cost of local computing scheme is highest, which is almost twice that of other algorithms. Meanwhile, the federated DDPG achieves 2\%$\sim$10\%, 9\%$\sim$20\% lower cost than federated DQN and F-AP computing algorithms respectively.

The average system costs of various algorithms with different CPU frequencies of F-APs are shown in Fig. 4. It is obvious that the average delay and energy costs of the federated DDPG, federated DQN and F-AP computing algorithms decrease as the CPU frequency of F-APs increases. Differently, the average system cost of local computing algorithm is almost constant because all tasks are executed without the use of F-AP resource. From the results, the system cost of federated DDPG algorithm is 9\%$\sim$25\%, 10\%$\sim$21\% lower than that of federated DQN and F-AP computing algorithms respectively. Therefore, the federated DDPG algorithm with the lowest system cost has the best performance under all CPU frequencies of F-APs.
\begin{figure}[t]
	\centering 
	\includegraphics[height=4.5cm,width=7.5cm]{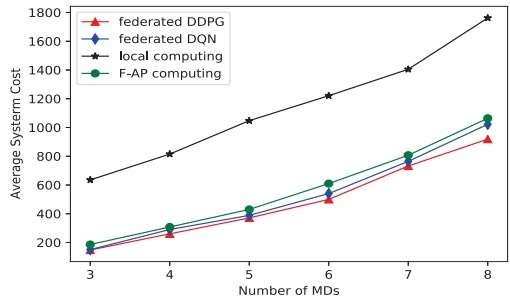}
	\caption{Average system cost versus number of MDs.}
	\label{2}
\end{figure}
\begin{figure}[t]
	\centering 
	\includegraphics[height=4.5cm,width=7.5cm]{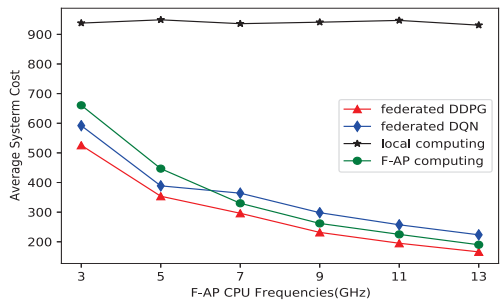}
	\caption{Average system cost versus F-AP CPU frequencies.}
	\label{3}
\end{figure}
\section{Conclusions}
In this paper, by considering the dynamic environment of the F-APs, the computation offloading and resource allocation problem with the goal of minimizing the user delay and energy consumption is studied. After casting the problem as an MDP, a federated DRL algorithm is proposed to adaptively learn the optimal policy of computation offloading and resource allocation, in which the DDPG algorithm is deployed on each F-AP and trained by the federated learning approach. Hence, the computing complexity is decreased and user privacy is protected. Simulation results demonstrate that the federated DDPG based computation offloading and resource allocation algorithm outperforms the corresponding centralized algorithm and value-based DRL algorithm, which converges faster to a higher system reward. Regardless of user density and F-AP CPU frequency, the federated DDPG algorithm always achieves the lowest system cost of task execution delay and energy consumption compared with the benchmark policies, proving the effectiveness of the proposed computation offloading and resource allocation scheme.

\section*{Acknowledgements}
This work was supported in part by the National Key Research and Development Program under Grant 2021YFB2900300, the National Natural Science Foundation of China under grant 61971129, and the Shenzhen Science and Technology Program under Grant KQTD20190929172545139.

\balance
\bibliography{reference}
\end{document}